# Electric Vehicle Driver Clustering using Statistical Model and Machine Learning

Yingqi Xiong, Bin Wang, Chi-Cheng Chu, Rajit Gadh
Mechanical and Aerospace Engineering
University of California, Los Angeles (UCLA)
Los Angeles, CA, USA
yxb936@ucla.edu

*Abstract*—Electric Vehicle (EV) is playing a significant role in the distribution energy management systems since the power consumption level of the EVs is much higher than the other regular home appliances. The randomness of the EV driver behaviors make the optimal charging or discharging scheduling even more difficult due to the uncertain charging session parameters. To minimize the impact of behavioral uncertainties, it is critical to develop effective methods to predict EV load for smart EV energy management. Using the EV smart charging infrastructures on UCLA campus and city of Santa Monica as testbeds, we have collected real-world datasets of EV charging behaviors, based on which we proposed an EV user modeling technique which combines statistical analysis and machine learning approaches. Specifically, unsupervised clustering algorithm, and multilayer perceptron are applied to historical charging record to make the day-ahead EV parking and load prediction. Experimental results with cross-validation show that our model can achieve good performance for charging control scheduling and online EV load forecasting.

## I. Introduction

The recent EV sales report [1] reveals that the population of EVs has increased tremendously in the past few years. Accordingly, the demand for installing charging stations in the electricity distribution system is also growing rapidly in order to meet drivers' travel demand. It is estimated that total number of EVs in U.S. by 2024 will reach 4 million [2]. A commercial EV charging station [3, p. 177] can deliver up to 120 Kw to EV batteries, which is equal to the aggregated load from 40 households [4]. However, uncontrolled EV charging behaviors may cause numerous issues for the power grids, including the increased operational cost, degraded power quality and the potential risk of power outage [5], [6].

To effectively manage the aggregated EV load, a number of approaches have been proposed by previous researches. For instance, Demand Response (DR) programs are offered to EV load aggregators to regulate the EV charging load according to the time-varying energy prices [7]. In addition, EV charging load can be deferred intelligently to different time windows, considering a number of different grid objectives, such as cost minimization [8]–[10], system load flattening and the valley-filling [11]. However, most of these approaches are based on assumptions that the charging session parameters, i.e. charging start time, leave time and energy consumption are pre-known without uncertainties, which is not realistic in most real-world cases. Thus, accurate session parameter prediction of the driver behaviors is needed by both smart charging and demand response programs. There are many well investigated forecasting methods for microgrid load, building load, solar generation [9], [12], [13], etc. However, there is lack of research works regarding EV driver clustering and EV load prediction. There are challenges to make day-ahead forecasting for EV load due to the following factors: 1) EV users are individuals with uncertain behaviors; 2) Due to the size of population, it is hard and not practical to model or label each EV user; 3) Load demand and capacity varies among different EV models.

Previous studies have partially covered some of those challenges discussed previously. A predictive EV charging control algorithm is proposed in [9], [14] where the randomness of user behavior is described by Kernel Density Estimation (KDE), which eliminates the restriction from specific distribution model, but no load forecasting has been performed. Uncertainty of EV user behaviors are also addressed in [15], where Markov Decision Process (MDP) and Queue Theory (QT) are utilized. [16] compares the performance of different EV load prediction methods including k-Nearest Neighbors (kNN) and Lazy-learning Algorithm. However, to the best of authors' knowledge, none of them provides a comprehensive solution to address the whole implementation cycle of such forecasting that can resolve all the aforementioned challenges.

This document was prepared as a result of work in part by grants from the California Energy Commission EPC-14-056 fund (Demonstration of PEV Smart Charging and Storage Supporting Grid Objectives Project). It does not necessarily represent the views of the Energy Commission, its employees, or the State of California. Neither the Commission, the State of California, nor the Commission's employees, contractors, nor subcontractors makes any warranty, express or implied, or assumes any legal liability for the information in this document; nor does any party represent that the use of this information will not infringe upon privately owned rights. This document has not been approved or disapproved by the Commission, nor has the Commission passed upon the accuracy of the information in this document.

In this paper, we utilize both supervised learning and unsupervised learning approaches to find patterns in EV user charging records and make prediction. K-means clustering is applied to categorize EV user behavior and neural network is used for further classification. The combination of both techniques is built for an online real-time EV user model to describe the uncertainty and eliminate the needs for hand labelling user behavior or revisiting historical dataset. The user model can be used by different kinds of scheduling algorithm for optimal predictive control scheme.

## II. SYSTEM OVERVIEW

### A. Smart EV Charging and Data Collection Infrastructure

The proposed work in this paper is developed using the UCLA SMERC smart charging network infrastructure as its testbed. Within this network there are more than 200 electric vehicle charging stations installed in public parking structures in multiple locations in Los Angeles region.

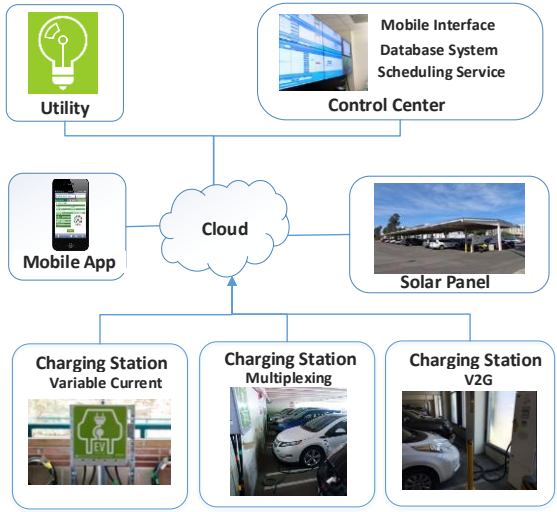

Figure 1. Smart EV charging infrastructure

Fig. 1. shows the architecture of SMERC smart charging infrastructure. This infrastructure has been designed and tested since 2013. There are mainly three layers of components in the system, i.e. the hardware level, communication network layer and the control center application layer. In the hardware layer, EV charger under SAE J1772 and CHadeMO are installed to provide energy to vehicle batteries. Within the control center, different energy management algorithms are implemented to regulate the EV charging behaviors considering different aspects of system properties, such as the solar generation, energy prices and the capacity of performing V2G operations, etc. Based on this system, EV drivers submit their preferences and monitoring the charging sessions, while the real-time monitoring data are collected and stored in the server-side databases.

### B. EV Driver Behavior Data

EV driver behavioral data has been collected for more than 4 years by the above-mentioned infrastructure. We collected the session parameters for each charging session, including the plug-in time, charging start time, charging stop time, charger plug-out, as well as charging current and power consumption values per minute. Several typical types of user records are shown in Fig. 2. to illustrate the statistical features of their behaviors.

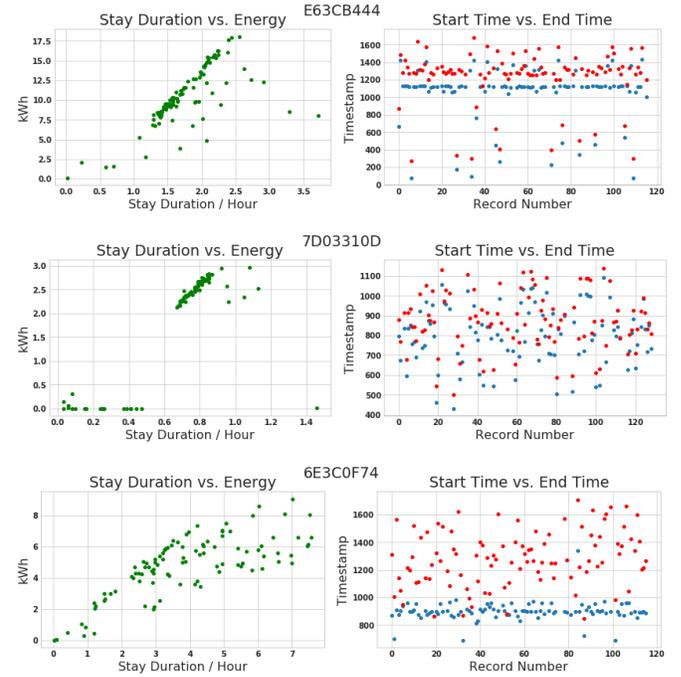

Figure 2. Some typical EV user behaviors

Three typical EV driver charging behaviors are displayed in Fig. 2. The correlations of EV stay durations and session energy consumptions are shown in the figures in the left column, while EV start charging time and session end time are displayed in the right column. From the visualization, one can find that it's difficult to have one universal distribution to model the behaviors of all EV drivers. Specifically, the charging behaviors of driver #E63CB444 are relatively easier to predict due to the roughly linear relationship between charging duration and energy demand. However, the driver #7D03310D has sparser arrival/end time and more stable energy consumption values. Interestingly, driver #6E3C0F74 has a very stable arrival time, but fluctuating end time. These characteristics represent different types of EV drivers, i.e. privately-owned EV drivers or fleet EV drivers, etc., whose charging behaviors may have different levels of impact on the distribution system and the effectiveness of smart charging programs. In this paper, we use unsupervised learning approach, i.e. K-Means, to capture these uncertainty characteristics of EV user behavior.

## III. EV USER BEHAVIOR MODELING AND ANALYSIS

### A. EV User Clustering

K-Means clustering is an unsupervised learning algorithm which can be used to partition data into k clusters based on some certain properties within each individual data point. We use this algorithm to process our historical charging records and generate assumptions of EV user behavior for EV charging scheduling.

From the visualization of historical EV charging record it can be observed that user behavior can be categorized into 4 groups. We choose the mean and standard deviation of arrival and departure time and the Pearson correlation coefficient between stay duration and energy consumption as clustering criteria. There are $k$ centroids $\mu_j \in R^5, j \in [1, k]$.

$$\mu := (\bar{t}_{arrival}, \bar{t}_{departure}, \sigma_{arrival}, \sigma_{departure}, cor) \quad (1)$$

where $cor$ is the Pearson correlation coefficient. The charging record data of each user are also processed into the same tuple structure as the clustering centroids. Assuming there are $m$ user records which give us the user clustering matrix $X \in R^{m \times k}$. The clustering algorithm updates the user group tag of each user in every iteration step by

$$c^i := \arg\min_j \|x^i - \mu_j\|^2 \quad (2)$$

where $c^i$ represents the user group tag of each user $x^i$, while $x^i \in X, i \in [1, m]$. Equation (2) changes the tag of each user to its closest user group. After updating all user group tags, the group centroid positions are then updated based on the users belonging by (3)

$$\mu_j := \frac{\sum_{i=1}^{m} I\{c^i = j\} x^i}{\sum_{i=1}^{m} I\{c^i = j\}} \quad (3)$$

The complete steps for EV user clustering are summarized in algorithm 1 as follows:

---
**Algorithm 1: EV User Behavior Clustering**

---
Process EV charging record for each user into tuple described in (1) to form the user clustering matrix $X$
Randomly initialize user group centroids $\mu_j, j = 1, 2, \dots, k$ from user clustering matrix $X$
    **While** $\sum_{i=1}^{m} \|x^i - \mu_{ji}\|^2 > \varepsilon$:
        **For** $i = 1, 2, 3, \dots m$:
            Calculate (2)
        **End**
        **For** $j = 1, 2, \dots, k$:
            Calculate (3)
        **End**
    **End**

---

The cost function selected for K-means algorithm is Euclidian distance. By minimize the cost function in (4) we can get the optimal group number:

$$\arg\min_j \sum_{i=1}^{m} \|x^i - \mu_{ji}\|^2 \quad (4)$$

### B. EV User Classification

Multilayer perceptron, also known as artificial neuron network, is a powerful tool to make classification over labelled dataset. It can capture features which are relatively hard to be observe by data visualization. In this paper we use multilayer perceptron to process EV user charging record data and make classification based on both clustering labels from K-means algorithm and hand-labelling by data visualization. The neuron network is trained using backpropagation and can be utilized for future classification on any new users without processing the whole dataset again. The structure of neuron network used in the proposed work is shown in Fig. 3.

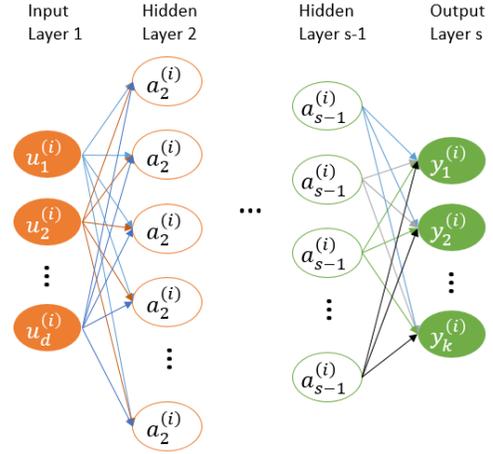

Figure 3. Multilayer perceptron network structure

For each user in the database charging records of d days are randomly picked to form the user charging record matrix $U \in R^{d \times m}$, the EV user number is denoted by $m$ for consistency, and $d$ is set up so that the feature length of each user is the same. User charging record matrix is transferred into the neuron network as input. The output of neuron network is a vector representing EV user groups. In the neuron network, each mapping from layer $s$-1 to layer $s$ follows the forward propagation rule:

$$a_s = g(\theta^{s-1} a_{s-1}) \quad (5)$$

Where $g(x)$ is the activation function which chosen to be sigmoid in this paper. $\theta^{s-1}$ is the weight matrix for mapping from layer $s$-1 to $s$. By forward propagation, features in the input user charging record matrix are extracted by each hidden layers and user behaviors are classified in the output vector.

Back propagation is used to train the neuron network. The output of user classification results are compared with labels from clustering algorithm and labels from visualization. Errors are computed for the output layer. Errors associated with output layer are then used to back-calculate the error with its preceding layer. The calculation is back propagated in the way until the input layer. Errors associated with each layer are used to calculate the partial derivatives to perform gradient descent to minimize the cost function:

$$J(\theta) = -\frac{1}{u} \left[ \sum_{i=1}^{u} \sum_{k}^{K} y_k^i \log\left(g(\theta a^i)\right) + (1 - y_k^i) \log\left(1 - g(\theta a^i)\right) \right] + \frac{\gamma}{2u} \sum_{s=1}^{s-1} \sum_{i=1}^{t_s} \sum_{j=1}^{t_{s+1}} (\theta_{ji}^s)^2 \quad (6)$$

Where *s* denotes the layer index and *t* is the number of neurons in that particular layer. The iteration goes on until convergence criteria is satisfied. The hyper-parameters, i.e. the number of hidden layers and the number of neurons in each layer is selected by random search and cross-validation, which is performed by first defining search ranges for each hyper-parameter, then randomly select parameters from search pools to train the neural network. The best hyper-parameter combination is recorded.

### C. EV User Model

From the user groups obtained by previous algorithms, we can generate the day-ahead predictive energy demand boundary. This demand boundary is used to construct the EV user model. We define there would be *N* EVs using charging stations within the network in the following day. Each EV has a charging/discharging rate $r_n(t)$ with $n \in N = [1, 2, ..., N]$ at time slot $t \in T = [1, 2, ..., T]$. *T* is the time span where the control can be performed.

The maximum charging rate and V2G rate of EVs made by different manufacturers are varied. We use $\overline{r_n}$ and $\underline{r_n}$ to denote maximum charging rate and maximum V2G rate of EV n, respectively. V2G rate limit $\underline{r_n}$ is a negative value since the power flow is reversed. Thus, we have the bi-directional charging rate limits for each EV:

$$\underline{r_n} \leq r_n(t) \leq \overline{r_n}, \text{ for } n \in N \text{ and } t \in T \quad (7)$$

Within the control algorithm implementation time span, not all EVs are online in every time slot. During the time when an EV is charging, the charging rate limits in (7) is effective. When an EV is not online, we set both the upper and lower bound of this EV charging rate to zero.

Energy consumption of each EV is denoted by $E_n$. It represents the energy demand of the particular EV during its stay in the parking facility. It is obvious that the summation of the products of EV charging rate and time interval equals to the total energy demand:

$$\sum_T r_n(t) \cdot \Delta t = E_n \quad (8)$$

where $\Delta t$ is a constant since the control time span is evenly divided into *T* sections.

According to the Central Limit Theorem (CLT), when the sample number reach some large value, the distribution of sample will converge to normal distribution. The availability time interval $[t_{arrival}, t_{departure}]$ of each EV can therefore be generated from clustering user groups using the mean and standard deviation. The energy consumption of EVs are then derived using the stay duration and the correlation in the user groups. We use a parameter $\beta_i, i \in [1, 2, ..., l]$ to define the portions of each user group to all EV users, and generate *l* charging rate boundaries from the *l* different user groups. The combined day-ahead EV user model is then:

$$R^{bound} = \sum_{l=1}^{l} \sum_{i=1}^{n} r_i^{bound} \beta_l \quad (9)$$

$$E^{total} = \sum_{l=1}^{l} \sum_{i=1}^{n} E_i \beta_l \quad (10)$$

### IV. RESULTS AND DISCUSSION

#### A. Experiment Setup

UCLA smart charging system as described in section II are used as testbed for the experiment. EV user charging records collected by the smart charging infrastructure for the past 4 years are used as the dataset for testing the proposed prediction algorithm. K-fold cross validation method is used to help trained the neuron network and selected hyper-parameters. In this paper we divide the dataset into 10 partitions, i.e. K=10. The descriptions dataset and partition setup are shown in table I and Fig. 4. respectively.

TABLE I
EV USER CHARGING RECORDS

|  | *Training Set* | *Test Set* |
|---|---|---|
| Number of Partitions | 9 | 1 |
| Number of Users | 110 | 20 |
| Number of Records | 11000 | 2000 |
| Number of Clusters | 4 | 4 |

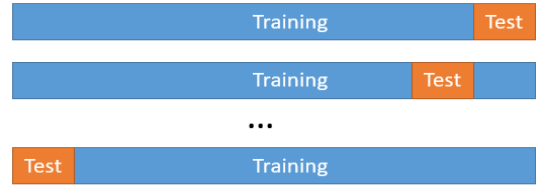

Figure 4. 10-fold cross validation

#### B. Prediction Performance Analysis

The clustering algorithm processes the historical EV user data in our database and divides EV user behaviors into 4 groups, as shown in Fig. 5. Three of the five features in the tuple are used for visualization with standard deviation normalized by mean. User group labeled by green has highly predictable behavior. Their arrival and departure schedule are fixed at certain timestamps with little variance. Their energy consumption is linear related with stay duration with Pearson score close to 1. On the contrary user group label in black has an almost random travelling schedule, using them as resources to participate in charging scheduling is highly unreliable.

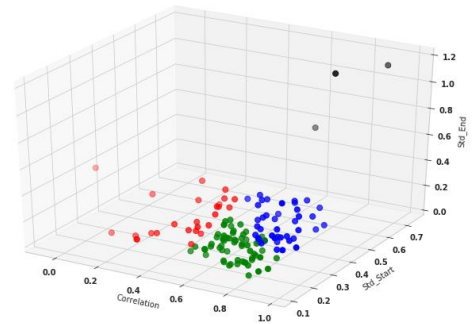

Figure 5. EV user behavior clusters

Multilayer perceptron neuron network is trained using collected EV user charging records and labels from both clustering algorithm and data visualization based hand-labelling. Cross validation is used to determine the hidden layer number and neurons per layer. According to the grid search results the optimal hidden layer number is 3 and neurons in each layer is 273, 212 and 169 respectively. By using the EV user model described in Section III part C, day-ahead EV user demand can be generated. The EV demand forecasting from clustering EV user model and from multilayer perceptron EV user model are plotted in Fig. 6. It can be seen that the deviation between the two curves is negligible.

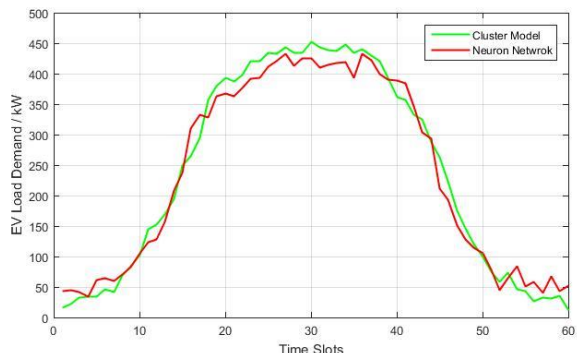

Figure 6. Day-ahead EV load demand from user models

More performance results are listed in Table II, where the neuron network classification accuracy with respect to the training set and test set. The neuron network model can achieve an average of 85% accuracy in the training set and around 78% in the test set. The EV load demand prediction is evaluated by Mean absolute percentage error (MAPE). It can be seen that the EV user model derived from multilayer perceptron is qualified for charging control scheduling with tolerable error.

TABLE II
MULTILAYER PERCEPTRON USER MODEL PERFORMANCE EVALUATION

| Cross Validation Case # | Classification Accuracy w.r.t Training Set | Classification Accuracy w.r.t Test Set | EV Load Demand MAPE |
|---|---|---|---|
| 1 | 82% | 78% | 0.254 |
| 2 | 76% | 72% | 0.351 |
| 3 | 79% | 74% | 0.312 |
| 4 | 91% | 82% | 0.145 |
| 5 | 88% | 84% | 0.176 |
| 6 | 87% | 73% | 0.221 |
| 7 | 83% | 72% | 0.213 |
| 8 | 85% | 75% | 0.267 |

## V. CONCLUSION

In this paper a data-driven method is introduced to model the EV user behavior based on EV user charging records collected in public charging facilities over 4 years. A novel approach which combines K-Means clustering and multilayer perceptron is developed and tested. The cross validation experimental results and performance evaluation show that proposed technique is capable to be used for various kinds of charging control scheduling. The proposed method makes labelling dataset become an automatic process and eliminates the need to perform clustering every time when a new user joins in the charging network and once trained, can be used parallel with real-time control.

REFERENCES

[1] "September EV Sales In US Hit 2017 High, And Its Only Up From Here!" [Online]. Available: https://insideevs.com/ev-sales-september-2017/. [Accessed: 08-Nov-2017].
[2] "Global Light Duty Electric Vehicle Sales Are Expected to Exceed Six Million in 2024," *Navigant Research*, 23-Dec-2015. [Online]. Available: https://www.navigantresearch.com/newsroom/global-light-duty-electric-vehicle-sales-are-expected-to-exceed-six-million-in-2024. [Accessed: 08-Nov-2017].
[3] "J1772: SAE Electric Vehicle and Plug in Hybrid Electric Vehicle Conductive Charge Coupler - SAE International." [Online]. Available: http://standards.sae.org/j1772_201210/. [Accessed: 08-Nov-2017].
[4] "How much electricity does an American home use? - FAQ - U.S. Energy Information Administration (EIA)." [Online]. Available: https://www.eia.gov/tools/faqs/faq.php?id=97&t=3. [Accessed: 08-Nov-2017].
[5] R. C. Green, L. Wang, M. Alam, and S. S. S. R. Depuru, "Evaluating the impact of Plug-in Hybrid Electric Vehicles on composite power system reliability," in *2011 North American Power Symposium*, 2011, pp. 1–7.
[6] B. Zhang and M. Kezunovic, "Impact of available electric vehicle battery power capacity on power system reliability," in *2013 IEEE Power Energy Society General Meeting*, 2013, pp. 1–5.
[7] B. Wang, B. Hu, C. Qiu, P. Chu, and R. Gadh, "EV charging algorithm implementation with user price preference," in *Innovative Smart Grid Technologies Conference (ISGT), 2015 IEEE Power Energy Society*, 2015, pp. 1–5.
[8] B. Wang, Y. Wang, C. Qiu, C. C. Chu, and R. Gadh, "Event-based electric vehicle scheduling considering random user behaviors," in *2015 IEEE International Conference on Smart Grid Communications (SmartGridComm)*, 2015, pp. 313–318.
[9] B. Wang, Y. Wang, H. Nazaripouya, C. Qiu, C. c Chu, and R. Gadh, "Predictive Scheduling Framework for Electric Vehicles Considering Uncertainties of User Behaviors," *IEEE Internet Things J.*, vol. PP, no. 99, pp. 1–1, 2016.
[10] Y. Wang, W. Shi, B. Wang, C.-C. Chu, and R. Gadh, "Optimal operation of stationary and mobile batteries in distribution grids," *Appl. Energy*, vol. 190, pp. 1289–1301, Mar. 2017.
[11] Y. Xiong, B. Wang, C. Chu, and R. Gadh, "Distributed Optimal Vehicle Grid Integration Strategy with User Behavior Prediction," *ArXiv170304552 Cs Math*, Mar. 2017.
[12] B. Wang *et al.*, "Predictive scheduling for Electric Vehicles considering uncertainty of load and user behaviors," in *2016 IEEE/PES Transmission and Distribution Conference and Exposition (T D)*, 2016, pp. 1–5.
[13] A. Gensler, J. Henze, B. Sick, and N. Raabe, "Deep Learning for solar power forecasting #x2014; An approach using AutoEncoder and LSTM Neural Networks," in *2016 IEEE International Conference on Systems, Man, and Cybernetics (SMC)*, 2016, pp. 002858–002865.
[14] Y. Xiong, B. Wang, Z. Cao, C. c Chu, H. Pota, and R. Gadh, "Extension of IEC61850 with smart EV charging," in *2016 IEEE Innovative Smart Grid Technologies - Asia (ISGT-Asia)*, 2016, pp. 294–299.
[15] T. Zhang, W. Chen, Z. Han, and Z. Cao, "Charging Scheduling of Electric Vehicles With Local Renewable Energy Under Uncertain Electric Vehicle Arrival and Grid Power Price," *IEEE Trans. Veh. Technol.*, vol. 63, no. 6, pp. 2600–2612, Jul. 2014.
[16] M. Majidpour, C. Qiu, P. Chu, R. Gadh, and H. R. Pota, "Fast Prediction for Sparse Time Series: Demand Forecast of EV Charging Stations for Cell Phone Applications," *IEEE Trans. Ind. Inform.*, vol. 11, no. 1, pp. 242–250, Feb. 2015.